\documentclass[conference]{IEEEtran}
\IEEEoverridecommandlockouts
\usepackage{cite}
\usepackage{amsmath,amssymb,amsfonts}
\usepackage{algorithmic}
\usepackage{graphicx}
\usepackage{textcomp}
\usepackage{xcolor}
\usepackage{multirow}
\usepackage{booktabs} 
\usepackage[ruled,vlined,linesnumbered]{algorithm2e}
\def\BibTeX{{\rm B\kern-.05em{\sc i\kern-.025em b}\kern-.08em
		T\kern-.1667em\lower.7ex\hbox{E}\kern-.125emX}}
\begin{document}
	
	\title{A PSO Based Method to Generate Actionable Counterfactuals 
		for High Dimensional Data
	}
	\author{\IEEEauthorblockN{Shashank Shekhar}
		
		\IEEEauthorblockA{\textit{Head - Global AI Practice} \\
			\textit{Subex Ltd,}\\
			Bengaluru, India \\
			shashank.shekhar@subex.com}
		\and
		\IEEEauthorblockN{Asif Salim}
		\IEEEauthorblockA{\textit{Research Scientist} \\
			\textit{Subex Ltd,}\\
		Bengaluru, India \\
		\hspace{2cm}asif.salim@subex.com\hspace{2cm}}
		\and
		\IEEEauthorblockN{Vivaswan Jinturkar}
		\IEEEauthorblockA{\textit{Senior Data Scientist} \\
			\textit{Subex Ltd,}\\
		Bengaluru, India \\
			\hspace{1cm}vivaswan.jinturkar@subex.com\hspace{1cm}}
			\and

		\IEEEauthorblockN{Anirudha Nayak}
		\IEEEauthorblockA{\textit{Senior Data Scientist} \\
			\textit{Subex Ltd,}\\
		Bengaluru, India \\
			 \hspace{2cm}anirudha.nayak@subex.com\hspace{2cm}
		}
		\and
		\IEEEauthorblockN{Adesh Bansode}
		\IEEEauthorblockA{\textit{Senior Data Scientist} \\
			\textit{Subex Ltd,}\\
		Bengaluru, India \\
			adesh.bansode@subex.com}
	
	}
	
	\maketitle
	
	\begin{abstract}
	Counterfactual explanations (CFE) are methods that explain a machine learning model by giving an alternate class prediction of a data point with some minimal changes in its features. It helps the users to identify their data attributes that caused an undesirable prediction like a loan or credit card rejection.
	We describe an efficient, and an actionable counterfactual (CF) generation method based on particle swarm optimization (PSO). We propose a simple objective function for the optimization of instance-centric CF generation problem. The PSO brings in a lot of flexibility in terms of carrying out multi-objective optimization in large dimensions, capability for multiple CF generation, and setting box constraints or immutability of data attributes. An algorithm is proposed that incorporates these features and it enables greater control over the \textit{proximity} and \textit{sparsity} properties over the generated CFs. The proposed algorithm is evaluated with a set of action-ability metrics in real-world datasets, and the results were superior compared to that of the state-of-the-arts. 
	\end{abstract}
	
	
	\section{Introduction}\label{intro}
	
	The explainable artificial intelligence (XAI) is an important area considering the wide adoption of industrial applications related to AI. This is especially important in the context of data regulations like GDPR \cite{voigt2017eu}. The commonly adopted tools for XAI are feature attribution (FA) methods and counterfactual explanations. The FA methods  \cite{lundberg2017unified}, \cite{ribeiro2016should}  correspond to those techniques that explain the model decision in terms of the input features. 
	
	But there are some sensitive applications where the users prefer to further investigate when a model decision turns unfavourable. In these situations, the users prefer to find the change in their input attributes for a favourable prediction. This is equivalent to the concept of counterfactual thinking (CFT) \cite{roese1997counterfactual} studied by psychologists in which
	human beings deduce the counter effects of an event. 
	
	Counter effects of a prediction can be critical in many data driven applications. A typical use case is when a black box machine learning model rejects the credit card application for a user. In this case, rather than knowing the feature attribution values, the users prefer for an actionable alternatives in the values of those features so that the model makes desired prediction. So CFE can be deployed in these cases where it is required to provide the  changes to be made to the data point where its original prediction may be changed to a desired class. 
	
	In general, CF explanations can be used to increase the trust of an automated decision-making system. Apart from this, CFs are used in classical statistics to make causal inference \cite{morgan2015counterfactuals}. Along this direction, CFs are also used to assess the fairness of a prediction model \cite{NIPS2017_a486cd07}. But this is out of scope of the work we are proposing and hence we limit our discussion to generating changes in attributes of a data point to change its model prediction to a desired class. In the next section, we discuss these concepts in detail and the significance of our proposal.
	
\section{Counterfactual Explanations} \label{cfe}

As per the definition provided by Molnar  \cite{molnar2019}, \textit{a counterfactual explanation of a prediction describes the smallest change to the feature values that changes the prediction to a predefined output.} It is an intuitive and user-friendly way of understanding a black-box machine learning model. The seminal work  that introduces the importance of CFE in the case of machine learning models was proposed by Wachter et.al \cite{wachter2017counterfactual}. The authors have listed three objectives of CFE that helps in interpreting a model prediction. 

\begin{enumerate}
	\item To help the user to understand why a particular decision was made by the model.
	\item To provide the inputs to the user in the case they want to challenge the model decision.
	\item To understand the changes to be made in the feature values so that the current prediction can be changed to a desired one.
\end{enumerate} 

These are particularly useful to explain the black box machine learning model predictions. In terms of the inner components or working principles of the learning algorithm, it may be difficult to give an explanation in the layman perspective.  This problem is further escalated when the decision function is non-linear and the data dimension is large. Hence to provide model explanations for a  layman, an efficient way is to show the changes in their data point and its effect in model prediction. This can also be characterized in multiple ways, for example, it is possible to give multiple numbers of CFE in which each of them has a different set of feature changes and it also possible to simulate "what-if" sort of situations. Hence the CFE is a great utility in the XAI landscape that interacts with the users in a simple way without getting to know about the inner working of algorithms and it also helps in generating trust between humans and automated decision systems. Apart from this, it helps the data scientists to keep their proprietary model details safe from public access.


For a robust CF generation there are several conditions to be satisfied. Some of these conditions have inter-dependencies while others are completely independent. Hence the underlying optimization process comes with a lot of challenges. To understand this, first we describe  the desired properties a CF generation algorithm is expected to have - interpretable and human friendly explanations, focusing on a small number of feature changes, avoiding contradictions, giving the predefined prediction as close as possible, feature values should be in the same range as the original distribution, and capability to generate multiple CFs that are diverse. To satisfy these properties, a CF generation technique shall ideally optimize the following quantities.

\begin{enumerate}
	\item \textit{Validity} - this quantity is defined to check whether the CFs are valid or if the class of the CF is the desired one.
	
	\item \textit{Proximity} - measures how close the CFs are to the data point queried.
	
	\item \textit{Sparsity} - this quantity refers to the number of feature changes in the CFs.
	
	\item \textit{Diversity} - this quantity is applicable in the case of multiple CF generation. In this situation, if the CFs are generated by the change in same set of features and the change in feature values are in same range, the CFs may not be actionable in real-world. 
	 
	\item \textit{Feasibility} - checks whether the CFs belong to the same distribution as of the dataset.
\end{enumerate}

The state-of-the-art methods are proposed to optimize a combination of these measures and they are discussed in Section \ref{prior_works}.

\subsection{Our motivation} \label{motivation}
There are two observations we found with state-of-the-arts that motivated our research. They are discussed below.

\subsubsection{To enable a simple multi-objective optimization using PSO} \label{pso_advantages}

The existing methods are formulated as a mutli-objective optimization problem with terms that have trade-offs and inter-dependencies. It is difficult to formulate the problem as a well-defined convex function optimization. Hence for practical use cases, the state-of-the-art methods  use either random changes to the feature values or a simple search around the neighborhood of the queried data point. For example, in the practical DiCE \cite{mothilal2020explaining} implementation that optimizes - \textit{validity}, \textit{proximity} and \textit{diversity} - the authors choose random feature change to generate CFs \cite{mothilal2020explaining}. Growing spheres (GS) \cite{laugel2017inverse} method searches the CFs in the hyper-sphere around the data point. Similarly, other state-of-the-arts probe into some specific property of CF generation with certain approaches. For example, FACE \cite{poyiadzi2020face} proposes a method in which the CFs are ensured to follow the underlying data distribution. For this purpose, they generate counterfactual from a "feasible path" with respect to the query point. The path is followed in the regions of significant density of data. 

We propose that a simple optimization procedure is the most practical way for real-world CF generation. As this  problem cannot be formulated as a convex optimization problem and there are inter-dependencies between the properties to be optimized, a complex optimization strategy with additional procedures may not serve the purpose. Also, in large dimensional data these complex procedures can create computation issues. Hence we put forward a simple PSO based optimization algorithm to generate CFs. There are many advantages in using PSO that help in easy CF generation as listed below. 


\begin{enumerate}
	\item \textit{Option to generate multiple CFs} - some state-of-the-art methods generate only one CF at a time. With PSO, we can have the subset of best performing particles to support multiple generation.
	
	\item \textit{Setting box constraints for the attributes} - Since PSO uses particle position update at each iteration, we can easily incorporate the range an attribute value can vary. This helps in enabling explicit control over \textit{proximity}.
	
	\item \textit{Setting immutable attributes} - For actionable CFs in the real world case, the users should have the option to set some attributes static. The PSO gives the option to avoid updates  for such attributes and it also helps in controlling \textit{sparsity} property.
	
	\item \textit{Handling categorical variables} - The common method of encoding of categorical variables is through one-hot encoding. With an advanced variant of PSO, we can have particle update that adhere to one-hot encoding. A detailed discussion is given in Section \ref{mdpso_motivation}. 
\end{enumerate}

\subsubsection{Analysis on high dimensional data}

The state-of-the-arts are always discussed on standard datasets on account of the convenience to interpret CFs. But those datasets are low dimensional. Examples are COMPAS, Adult-income, German-credit and LendingClub. The studies on CF generation on high dimensional data are scarce. Hence we choose high dimensional datasets for our study with COMPAS included as a standard. 

The PSO provides default settings to make faster optimization in terms of runtime. This can be achieved by parallel updation of particle positions. Hence the proposed approach based on PSO is suitable for high dimensional data.

\section{Related Works} \label{prior_works}

Following the taxonomy described by Chou et.al \cite{chou2022counterfactuals}, our work is more interested in instance centric methods. These methods are the ones that seek CF based on feature changes to propose an instance that is close to the queried point. They are briefly discussed below.

The seminal work on CFE is done by Wachter et.al \cite{wachter2017counterfactual}. They proposed a base method in which CFs are generated considering only the optimization of \textit{proximity}. DiCE is a popular method proposed by Mothilal et.al  \cite{mothilal2020explaining} . 
They devised a method to generate diverse CFs by formulating a mutli-objective optimization that takes into account the properties of \textit{proximity}, \textit{sparsity} and \textit{diversity}. Dandl et.al \cite{dandl2020multi} devised CF generation via a multi-objective optimization that optimizes the previous three along with \textit{feasibility}. Looveren et.al  \cite{looveren2021interpretable}  used the prototypes of data points using autoencoder or class specific k-d trees. The prototypes are used in optimizing an objective function to ensure the generated CFs are \textit{feasible}. Along with this, the \textit{proximity} and \textit{sparsity} components are also optimized. 

Poyiadzi et.al \cite{poyiadzi2020face} proposed a method for feasible and actionable CF generation (FACE). Those properties are enabled by ensuring the CFs to lie in a high density region of data points. They claim that with this feature, the generated CFs are close to the data manifold and hence feasible. Growing spheres proposed by Laugel et.al \cite{laugel2017inverse} is an intuitive search algorithm to find a CF in the hypersphere centered around the queried data point.  The CF can be an actual data point in the dataset or a sampled point in the sphere.

The methods so far discussed are model agnostic. Apart from this, Artelt et.al  \cite{artelt2019computation} proposed model-specific  CF generation techniques for a collection of algorithms such as linear models, Gaussian naive Bayes, quadratic discriminant analysis etc. 

Another class of CF generation methods are named as probabilistic methods. They use the techniques based on probabilistic graphical models or generative models such as variational auto-encoder. The examples are CRUDS \cite{downs2020cruds}, CEM \cite{dhurandhar2018explanations}, CLUE \cite{antoran2021getting}, and REVISE \cite{joshi2019towards}.
\section{Proposed Method} \label{method}

In this section, we describe the PSO variant that is used for our modeling, the objective function optimized for CF generation, the evaluation metrics for checking the efficacy of the generated counterfactuals and the CF generation algorithm.

\subsection{The PSO variant}
We have used a mixed discrete particle swarm optimization algorithm (MD-PSO) \cite{chowdhury2013mixed} in our experiments. The velocity updation equation in MD-PSO can be written as follows.

\begin{equation}
	V_i^{t+1} = c_0 V_i^t + c_1r_1 (P_i - X_i^t) + c_2r_2(P_g - X_i^t) + c_3r_3\hat{V}_i^t
\end{equation}

Particle update equation is,

\begin{equation}
	X_i^{t+1} = X_i^t + V_i^{t+1}
\end{equation}

where,
\begin{itemize}
	\item  $i$ denotes the particle:$i$, $t$ denotes the iteration:$t$,
	\item $X_i^{t}/X_i^{t+1}$ denotes the particle position at iteration $t$ and $t+1$ respectively,
	\item  $V_i^{t}/V_i^{t+1}$ denotes the velocity of the particle at iteration $t$ and $t+1$ respectively,
	
	\item $P_i$ is the local best position of particle:$i$ till iteration $t$,
	\item $P_g$ is the global best position among all particles till iteration $t$,
	\item $c_0$, $c_1$, and $c_2$ are hyper-parameters that determine the properties of inertia, exploitation, and exploration properties of the particles,
	
	\item $r_1$, $r_2$, and $r_3$ are real random numbers between [0, $c_1$], [0, $c_2$], and [$c_1$, $c_2$] respectively,
	
	\item $c_3$ is the diversity preservation coefficient proposed by the authors for continuous variables,
	\item The term $c_3r_3\hat{V}_i^t$ is explicitly introduced by the authors to ensure diversity among the particles. $\hat{V}_i^t$ is characterized as a vector with a divergence property. An example is  the vector connecting the mean of the population of particles (or its best performing subset) to the particle of interest.The purpose of introduction of this component is to avoid the stagnation of the particles at local minima. 
\end{itemize}

\subsection{Motivation to choose MD-PSO for CF generation} \label{mdpso_motivation}

As explained in section \ref{motivation}, PSO gives some default advantages that are required for a CF generation. Adding to this, PSO provides default parallel computing capabilities for optimization in large dimensions. Apart from giving the options to generate multiple CFs and setting box constraints, MD-PSO brings in two features that are critical in CF generation process as explained below.

\begin{enumerate}
	\item Support for generating diverse CFs.
	
	In an ideal PSO convergence scenario, all the particles are supposed to share the position around a small region. Hence if we take the subset of best performing particles in terms of minimizing the objective function, they may be close points to each other. Hence from a CF generation perspective,  the fourth term introduced in MDPSO provides an option to enhance diversity. 
	
	The straight forward way to pick CFs from the PSO output is to choose a subset of best performing particles. To further ensure diversity, we will do a clustering on the particles and pick those ones from the cluster that have a predefined support for the action-ability metrics as explained in Section \ref{selection}. We assume that the fourth term in MD-PSO helps in creating meaningful clusters.
	
	\item Easy integration of discrete variables in the optimization process.
	
	In analogous to the continuous variable updations in basic variant of PSO, MD-PSO maps the discrete variables to a hypercube \cite{chowdhury2013mixed}. Similar to the variables in continuous domain, the discrete variables are allowed to take real values confined to the hypercube. During updation process, they are mapped to the nearest corner of the hypercube to restore their discrete property.
	
	In real-world scenarios, the discrete variables, for example - gender, race, nationality, etc can turn crucial for a CF analysis. Hence it is important to integrate these discrete variables with the continuous ones and MD-PSO facilitates this.
	
\end{enumerate}

\subsection{Objective function}

Inspired from \cite{dandl2020multi}, we use the following objective function  for the CF generation,

\begin{equation}\label{obj}
	l_1\left( f(x_{cf}), y_{cf}\right) + 	l_2\left( x_0, x_{cf} \right) + l_3\left( x_0, x_{cf} \right)
\end{equation}

where $x_{cf}$ is the counterfactual instance of a data point of interest denoted by $x_0$, $y_{cf}$ is the desired class as against that of $x_0$, and $f(.)$ is a classifier. The individual components $l_1$, $l_2$, and $l_3$ are defined as follows.

\begin{equation}
	l_1\left( f(x_{cf}), y_{cf}\right) = 
	\begin{cases}
		0 & \text{if $f(x_{cf}) \in y_{cf}$}\\
		\underset{y_{cf}}{\text{inf}} |f(x_{cf}) - y_{cf}| &  \text{else}
	\end{cases}       
\end{equation}

The function $l_1$ is introduced to make the prediction of CF as close as possible to the desired prediction.  

\begin{equation}
	l_2\left( x_0, x_{cf} \right) = \frac{1}{p} \sum_{j=1}^{p} \delta_G(x_0^j, x_{cf}^j)
\end{equation}

where $p$ is the data dimension and $\delta_G$ is the Gower distance which is defined as,

\begin{equation}
	\delta_G(x_0^j, x_{cf}^j) = 
	\begin{cases}
		\frac{1}{R_j} |x_0^j - x_{cf}^j|& \text{if $x_0^j$ is numeral}\\
		\textbf{1}_{x_0^j \neq x_{cf}^j} & \text{if $x_0^j$ is categorical}
	\end{cases}       
\end{equation}

Here $R_j$ is a normalizer to scale the feature values. This term ensures that the counterfactual instance should be as close as possible to the data point of interest as this property is crucial for its action-ability. 

\begin{equation}
	l_3\left( x_0, x_{cf} \right) = \Vert x_0 - x_{cf} \Vert_0 = \sum_{j=1}^{p} \textbf{1}_{x_0^j \neq x_{cf}^j}
\end{equation}

where $\textbf{1}$ denotes the instance when the sub-scripted inequality is satisfied. The purpose of this term is to bring in minimal feature changes to generate actionable CF. The objective function is minimized using MD-PSO to generate a diverse set of CFs.

The mathematical form of $l_2$ and $l_3$ may look same but they serve entirely different purposes. The CF is expected to be similar to the query point for its action-ability and hence $l_2$ tries to minimize the \textit{proximity} property. On the other hand, it is also desired to keep the number of features changes in the generated CF to be minimum with respect to the query point. Hence adding  $l_3$ to the objective function minimize the \textit{sparsity} property that has to be satisfied by a CF. Suppose if a CF is generated with low sparsity but with a large number of features changes, such a case may be not actionable in the real world. Hence we use $l_2$ and $l_3$ in the objective function optimization to ensure optimal \textit{proximity} with minimal \textit{sparsity}.

The remaining properties that has to be satisfied for faithful CF generation are \textit{validity}, \textit{diversity}, and \textit{feasibility}. \textit{Validity} can be easily ensured as it is a necessary requirement for a data point to be qualified as a CF. Our proposed algorithm supports multiple CF generation by default. The \textit{diversity} property among those are ensured by adding a new term to the velocity update of the particle in the MD-PSO as explained in Section \ref{mdpso_motivation}. Among instance centric methods, only FACE ensures \textit{feasibility} while the probabilistic methods can  ensure this. But this is enabled by costly methods that involve model training  of generative algorithms. Since the scope of our study is only limited to instance centric methods, any specific component for \textit{feasibility} is not added. However, we propose that the MD-PSO objective formulation along with the control over \textit{proximity} and \textit{sparsity} (as explained in Section \ref{motivation}) ensures CF generations that are \textit{feasible}.

\subsection{Quality evaluation of conterfactuals} \label{quality_metrics}
The counterfactuals need to be evaluated to check if it is effective. Towards this direction, there are multiple quality checks that can be done in terms of their \textit{validity}, \textit{proximity}, \textit{sparsity}, and \textit{diversity}. These measures are also closely related to the individual terms we are trying to optimize in the objective function given by   \eqref{obj}. 

It has to be noted that there are no standardized way to evaluate the CFs in the research community and hence we use these metrics inspired from Mothilal et.al \cite{mothilal2020explaining}. The quality evaluation metrics used in our study are explained as follows.

\begin{enumerate}

	\item Proximity: This quantity measures how close the generated CFs are to the data point of interest. For the real-world users, it is important to get the CFs that are close to their original data instance. The measure is defined separately for continuous and categorical variable as follows.
	
	\begin{equation} \label{prox_cont}
		\text{proximity}_{cont} := \frac{1}{k} \sum_{i=1}^{k} 	d_{cont}(x_{cf}^i, x_0)
	\end{equation} 
	
	where $d_{cont}$ is defined as,
	
	\begin{equation} \label{dist_cont}
		d_{cont}(x_{cf},x_0) = \frac{1}{n_{cont}}  \sum_{j=1}^{n_{cont}} \frac{\left|x_{cf}^{[j]} - x_{0}^{[j]}\right|}{MAD_j}
	\end{equation}
	Here, $k$ denotes number of CFs generated, superscript $[j]$ denotes the $j^{th}$ component of $x_{cf}$ and $x_0$, $n_{cont}$ denotes the number of continuous variables and $MAD_j$ stands for the \textit{median absolute deviation} of the  $j^{th}$ component.
	
	The proximity measure for categorical variables can be defined as,
	
	\begin{equation}\label{prox_cat}
		\text{proximity}_{cat} :=  \frac{1}{k}\sum_{i=1}^{k} d_{cat}(x_{cf}^i, x_0)
	\end{equation}
	where $d_{cat}$ is defined as,
	
	\begin{equation} \label{dist_cat}
		d_{cat}(x_{cf},x_0) = \frac{1}{n_{cat}} \sum_{j=1}^{n_{cat}} \textbf{1} (x_{cf}^{[j]} \neq x_{0}^{[j]}) 
	\end{equation}
	
	and $n_{cat}$ denotes the number of categorical variables. The optimal CF generation requires this metric to be low.
	
	Note that \eqref{prox_cont} and \eqref{prox_cat} can be used in the place of $l_2$ in the PSO objective function given in \eqref{obj}. However, $l_2$ is used as it is computationally efficient.
	\newline
	\item Sparsity:  measures the number of attributes that are changed with respect to $x_0$. For actionable CFs, the users expect minimal action and hence minimal changes to the number of features. The measure is defined as follows,
	
	\begin{equation}
		\text{sparsity} :=  \frac{1}{p}\sum_{i=1}^{p} d_{spa}(x_{cf}^i, x_0)
	\end{equation}

where $d_{spa}$ is defined as,
	\begin{equation}
		d_{spa}(x_{cf},x_0) = \frac{1}{p}  \sum_{j=1}^{p} \textbf{1} (x_{cf}^{[j]} \neq x_{0}^{[j]}) 
	\end{equation}
	
	where $p$ is the total number of features or attributes. This quantity is expected to be low for actionable CFs.
	\newline
		\item Normalized Diversity: as the name suggests it is a measure to check how different the generated CFs are compared to each other. For actionable CFs, we need to generate the ones that would not contradict with each other and at the same time keeps the close proximity to the data point of interest. Here the contradiction means that two CFs should not have opposite feature attributions, that is, they shall not recommend a feature to be increased and decreased for the same query point.  They shall also focus on minimal changes to the features and at the same time shall be diverse enough to give the users multiple options for their actions. These is a  requirement with trade-offs and considering this we define the metric diversity as a distance metric between each pair of CF examples as shown in \eqref{diversity}.
	
	\begin{equation}\label{diversity}
		\text{diversity} := \frac{1}{kC2} \sum_{i=1}^{k-1} \sum_{j=i+1}^{k} d(x_{cf}^i, x_{cf}^j)
	\end{equation}
	
	where we assume $k$ number of CFs are generated, $x_{cf}^i$ denotes the $i^{th}$ CF and $d()$ is a distance metric that can be defined separately for continuous (as in \eqref{dist_cont}) and categorical variables (as in \eqref{dist_cat}). 
	
	The expected range of this term for actionable CFs is relative in nature. A low value indicates that CFs are similar to each other, and high value indicates that CFs are different from each other. This situation may also affects the action-ability of generated CFs. This quantity is also influenced by \textit{sparsity} as its high value indicates that a large number of features may be changed and hence it may trigger a larger \textit{diversity} value which is not desirable. For this reason, as shown in \eqref{diversity_norm}, we use \textit{normalized diversity} score in which normalization is done by the dividing \eqref{diversity} by \textit{sparsity} value.
	
		\begin{equation}\label{diversity_norm}
		\text{diversity$_{nrm}$} := \frac{1}{kC2} \sum_{i=1}^{k-1} \sum_{j=i+1}^{k} \frac{d(x_{cf}^i, x_{cf}^j)}{\text{prox}(x_{cf}^i) + \text{prox}(x_{cf}^j)}
	\end{equation}

where $\text{prox}(x_{cf}^i),\;\text{prox}(x_{cf}^j)$ indicates the \textit{proximity} of the CFs $x_{cf}^i,\;x_{cf}^j$ with respect to the query point, $x_0$.
	\newline
	\item Coverage: this quantity stands for the fraction of valid CFs. In our experiments, we generate multiple CFs and this measure counts how many instances the CF is a valid one. We consider a CF generation output invalid if it generates CF belonging to the same class as the query point or if the generation method was not able to generate any output at all. This measure is expected to be maximum for a CF generation method.
\end{enumerate}

\subsection{MD-PSO algorithm for counterfactual generation} \label{mdpso-algo}
In this section, we describe our algorithm and how we incorporated the advantages mentioned in Section \ref{pso_advantages}. 

For actionable CF generation, we have formulated an algorithm that gives a control over the \textit{proximity} and \textit{sparsity} evaluation metrics. The idea is to limit the number of features that can change and this feature subset given the name \textit{active set} ($\mathcal{A}$), is selected based on the \textit{information value (IV)} statistic. The number of features that can be changed or cardinality of \textit{active set} is set as a hyper-parameter, $h$. In other words, the top $h$ features are selected after arranging them in descending order of \textit{IV}. This helps in controlling the \textit{sparsity} metric. A description of this statistic is provided in Appendix. 

 As the next step, box constraints are set to the features in $\mathcal{A}$. We scale the dataset in the interval [0,1] for the experiments. The constraints are enabled in the form of a hyper-parameter, $\epsilon$, that characterizes the range features can vary and it enables control over \textit{proximity}. Note that the box constraint is enabled in the form of $[z(1-\epsilon), z(1+\epsilon)]$ if $z$ is the feature value. If the interval bounds falls outside [0,1] they will be capped to the extremes of 0 or 1. The processes are summarized in Algorithm \ref{alg1}.

\begin{algorithm}
	\label{alg1}
	\DontPrintSemicolon
	\SetAlgoLined
	\SetKwInOut{Input}{Input}\SetKwInOut{Output}{Output}
	\Input{decision function $f(.)$, query data point $x_0$, training dataset $D$, box constraint parameter $\epsilon$, number of mutable features $h$}
	\Output{A set of counterfactuals}
	\BlankLine
	
	\For{each feature in $D$}{
		Find the information value \;
	}
	
	Arrange the features in the decreasing order of $IV$ \;
	Select the top $h$ features to form $\mathcal{A}$ \;
	
Initialize the particle values satisfying $\epsilon$-criterion with respect to $x_0$ \;

Run MDPSO algorithm with the features defined in $\mathcal{A}$\;

Apply CF selection criteria on particles \;
	
	Return selected CFs \;
	\caption{MDPSO algorithm for counterfactual generation}
\end{algorithm}

Note that in our experiments, the parameter $\epsilon$ is common for all features. However, based on the domain knowledge custom parameters can be used for different features.  The CF selection criteria in line-8 of Algorithm \ref{alg1} is explained in Section \ref{selection}.
\section{Experiments}

In this section, we describe the experiment settings, datasets used, results and related observations.

\subsection{Datasets}

To check the effectiveness of the proposed approach, we used three high dimensional datasets namely \cite{dressel2018accuracy}, Arrhythmia \cite{arrhythmia}, hill-valley \cite{hill-valley}, and LSVT \cite{lsvt} along with the standard,  COMPAS dataset.

COMPAS dataset is the details of 10,000 persons with criminal charges in the state of Florida. The dataset was used to grant parole by checking  into the likelihood of re-offending the crimes. 
The Hill-valley dataset consists of a collection 100 points in a 2D graph that takes the form of a hill or a valley. 
The Arrhythmia dataset is a collection of patient records to detect the presence and absence of cardiac arrhythmia. The original dataset is of 16 classes. In our experiments, we converted them into a binary classification problem by taking the 'normal' class and rest of the classes combined as the other class. The LSVT is a dataset that  assess whether voice rehabilitation treatment lead to phonations considered 'acceptable' or 'unacceptable' in patients suffering from Parkinson's disease. The data points consist of a set of features from voice processing algorithms.
 
\begin{table}
	\caption{Dataset details. Size indicates the number of data points, Cont.feat - number of continous features and Cat.feat - number of categorical features.}
	\label{table1}
	\centering
	\scalebox{1}{
		\begin{tabular}{llllll}
			
			\toprule
			Dataset    & Classes & Size &  Cont.feat. & Cat.feat.\\ 
			\midrule

			COMPAS & 2 & 5238 &	13 & 0 \\ 
			
			Hill-Valley & 2 &  1212 & 100 & 0 \\ 
			
			Arrhythmia & 2 & 452 &	177 &	85 \\ 

			LSVT & 2 & 126 & 304 & 3\\ 
			
			\bottomrule
	\end{tabular}}
\end{table}

\subsection{Experimental setup}

In this section, we describe the parameters related to the models we have used for the experiments and other design parameters.

\subsubsection{General settings}
For PSO, we have used 50 number of particles and the total number of iterations or generations are set to 100. The parameters $c_i, i=\{0,1,2,3\}$ and 
$r_i, i=\{1,2,3\}$ are fixed by cross-validation.

The number of data points whose CFs need to be determined are taken as 5 and they are chosen at random from the testing set. We generate 5 number of CFs for each of these data points. The metrics used to evaluate the quality of CFs are the ones explained in Section \ref{quality_metrics}. The results reported are averaged across each instance of the CF generation.

The model chosen for classification is selected as the best performing one among logistic regression, artificial neural networks and random forest. Once the model for a dataset is fixed, it is saved and consumed by the CF generation methods. Note that in our experiments, random forest was the best performing model for all datasets. Note that in this study, the selection of classifier is merely an enabler to supply the decision function $f(.)$.

\begin{table*}[]
	\caption{Results of CF generation experiments.}
	\label{results}
	\centering
	\begin{tabular}{clllllc}
		\hline
		\textbf{Dataset}              & \multicolumn{1}{c}{\textbf{Method}} & \multicolumn{1}{c}{\textbf{Proximity}} & \multicolumn{1}{c}{\textbf{Sparsity}} & \multicolumn{1}{c}{\textbf{Diversity}} &
        \multicolumn{1}{c}{\textbf{Norm. Div}} &
		 \multicolumn{1}{c}{\textbf{Coverage}} \\
		\hline
		& DiCE-Random                         & 1.3992                                 & 0.1999                                & 1.7163  & 1.2266                               & \textbf{1.00}                                              \\
		& DiCE-Genetic                        & 1.5537                                 & 0.3571                                & 0.0437    & 0.0281                             & \textbf{1.00}                                         \\
		& GS                                  & 1.1109                                 & 0.4464                                & 1.9586              & 1.7630                   & 0.80                                           \\
		& Face-KNN                            & 1.5928                                 & 0.3714                                & 1.8357          & 1.1525                       & \textbf{1.00}                                        \\
		& Face-Epsilon                        & 1.6795                                 & 0.3828                                & 1.8725       & 1.1149                          & \textbf{1.00}                                        \\
		\multirow{-6}{*}{COMPAS}      & MD-PSO                                 & \textbf{0.2629}                                 & \textbf{0.1230}        &  0.2959   & 1.1257     & \textbf{1.00}         \\
		\hline

		& DiCE-Random                         & 0.9079                                 & \textbf{0.0633}                               & 1.0800     & 1.1895                              & \textbf{1.00}                                             \\
		& DiCE-Genetic                        & 0.5047                                 & 0.7227                                & 0.6890                & 1.3651                  & 0.80                                            \\
		& GS                                  & 1.0040                                  & 1.0000                             & 0.9864 & 0.9824                                 & 0.60                                         \\
		& Face-KNN                            & 1.1806                                 & 0.7980                                 & 0.9885  & 0.8372                               & \textbf{1.00}                                           \\
		& Face-Epsilon                        & 1.0022                                 & 0.7976                                & 0.8417        & 0.8398                         & \textbf{1.00}                                            \\
		\multirow{-6}{*}{Hill-valley}      & MD-PSO                                 & \textbf{0.0700}                                 & 0.6440        & 0.0571          & 0.8172 &    \textbf{1.00}                    \\
		
		\hline
		& DiCE-Random                         & 1.6771                                 & 0.2159                                & 2.9967       & 1.7868                          & \textbf{1.00}                                             \\
		& DiCE-Genetic                        & 2.3337                                  & 0.5269                                & 2.3836 & 1.0213                                & \textbf{1.00}                                        \\
		& GS                                  & 1.1168                                 & 0.9962                                & 2.1829     & 1.9546                            & 0.60                                         \\
		& Face-KNN                            & 2.0494                                 & 0.4188                                & 2.6015      & 1.2693                           & \textbf{1.00}                                        \\
		& Face-Epsilon                        & 2.6416                                 & 0.5075                                & 2.4496       & 0.9273                          & \textbf{1.00}                                             \\
		\multirow{-6}{*}{Arrhythmia}      & MD-PSO                                 & \textbf{0.9989}                                 & \textbf{0.0865}        & 1.0452          &   1.0464 & 0.60                       \\
		
		\hline

		& DiCE-Random                         & 1.4819                                 & \textbf{0.1639}                                & 2.7050   & 1.8253                               & 0.80                                              \\
		& DiCE-Genetic                        & 2.1654                                       &  0.8665                                     &  2.3250      & 1.0737                                & 0.80                                                 \\
		& GS                                  & \textbf{1.0056}                                 & 0.9967                                & 1.6997    & 1.6902                             & 0.60                                             \\
		& Face-KNN                            & 1.7700                                   & 0.7759                                & 2.1967    & 1.2410                             & \textbf{1.00}                                             \\
		& Face-Epsilon                        & 2.3219                                 & 0.9713                                & 2.1201      & 0.9130                           & \textbf{1.00}                                          \\
		\multirow{-6}{*}{LSVT}      & MD-PSO                                 & 1.7859                               & 0.3933        & 0.9527 & 0.5334          &     0.80                   \\
		\hline      
	\end{tabular}
\end{table*}

\subsubsection{Counterfactual selection criteria}
\label{selection}

After the PSO convergence, we have multiple options to select the $k$ number of CFs. The basic method is to choose the top $k$ best particles in terms of minimizing the objective function. But if the PSO convergence is proper, there is a high probability that the best $k$ particles are close to each other. Hence it can adversely affect the \textit{diversity} metric.

Another option is to keep track of the individual best positions of each particle across all iterations. The problem with this approach is that it can create additional computation overhead and mostly these individual bests might have occurred at the last set of iterations.

To ensure a balance between the evaluation metrics, we have done a k-means clustering over the converged particle positions and a CF is randomly chosen from each cluster. The number of clusters is taken as the number of CFs intended to be created which is 5 in our experiments. We have found that this third approach was giving optimal results compared to the first two.

\subsubsection{Settings of state-of-the-arts}

The scope of our study is limited to only instance-centric methods and hence the proposed method is compared with the following state-of-the-arts - growing spheres (GS) \cite{laugel2017inverse}, DiCE \cite{mothilal2020explaining} with \textit{random} and \textit{genetic algorithm} variants and FACE \cite{poyiadzi2020face} with \textit{EPS} and \textit{KNN} variants. Among these methods only DiCE supports multiple CF generation. For other cases, we run them 5 times with different initializations to get multiple CFs.

The GS and FACE methods are implemented with the CARLA library \cite{pawelczyk2021carla} and DiCE with the library published by authors \cite{mothilal2020explaining}. For GS, the step size value for growing the sphere is chosen as 0.02. In the case of FACE,  hyperparameters for the graph size is tuned by a grid search \cite{9718485}. The value of radius for the epsilon graph is chosen as 0.25 after cross-validation.

\subsection{Results and Observations}


The results obtained in our experiments are tabulated in Table \ref{results}. The best results are bolded. Note that the desired values of \textit{proximity} and \textit{sparsity} is to be minimum and that of \textit{coverage} is maximum. However, it is difficult to set the desired range for both the \textit{diversity} metrics as explained in Section \ref{quality_metrics}. In this case, for the sake of checking the quality of a method, we can check whether the scores of the method lies between the minimum and maximum scores among all methods. If it is so, that particular method would not be generating CFs that have too low or high values of diversity. 

From the results, it can be seen that proposed MD-PSO method has the best \textit{proximity score} in all datasets except LSVT. In the case of LSVT, the best result is with GS but its \textit{sparsity} and \textit{normalized diversity} score is very high. 

In terms of \textit{sparsity}, MD-PSO has the best results in COMPAS and Arrhythmia. In the case of Hill-valley and LSVT datasets, MD-PSO has the second best result following DiCE random. However, in Hill-valley,  DiCE Random has a  high \textit{proximity} score (0.9079) compared to MD-PSO (0.07). 

The \textit{coverage} results of MD-PSO is maximum in the case of COMPAS and Hill-valley datasets and the second best in other datasets. The FACE variants are the only models that managed to cover the complete requirement of 5 number of CF generations for all the query points. However, these models have the highest \textit{proximity} score which affects the action-ability of CFs. 

It can be seen that none of the models are able to give the best results across all datasets. However, MD-PSO is having consistent performance in the metrics relative to the state-of-the-arts. The results also validate the effectiveness of the control mechanism over the \textit{proximity} and \textit{sparsity} scores in the proposed algorithm as explained in Section \ref{mdpso-algo}.

Note that both the \textit{diversity} and \textit{normalized diversity}  is provided in the results. This is given to understand the inter-dependence between \textit{proximity} and \textit{diversity}. This dependency is more evident in the following example - in the case of MD-PSO result in Hill-valley, the \textit{diversity} score (0.0571) is the lowest among all methods due to a low \textit{proximity} score. However, the \textit{normalized diversity} score (0.8172) is in par with the rest of the methods. Hence \textit{normalized diversity} score helps in making meaningful comparison of the diversity of the generated CFs among the methods whose \textit{proximity} score varies in a large range. 

It can be noted that the  \textit{normalized diversity} score of MD-PSO lies between the maximum and minimum in the case of COMPAS, Hill-valley and Arrhythmia datasets. Hence the divergence levels among the CFs lie between the extremes and indicate the effectiveness of the proposed approach.

\section{Conclusions}

We designed an algorithm for counterfactual generation in high dimensional datasets based on particle swarm optimization. The objective function is formulated as a multi-objective function comprising the properties that influence actionable generation, namely, \textit{validity}, \textit{proximity}, and \textit{sparsity}. We adopted the MD-PSO in which an explicit control is enabled over \textit{proximity}, and \textit{sparsity} properties and suitable to optimize categorical attributes in the one-hot encoded format. With this features, we proposed an algorithm that helps in CF generations that are actionable in the real-world. The method was tested on real-world datasets and the results are found to be superior compared to that of the state-of-the-arts. As future work, we consider the optimization problem of CF generation in the context of Bayesian optimization.

\appendix 

Information value (IV) is a measure that helps to identify how a predictor affects in the prediction of target variable. To calculate this value, we divide each feature into a finite number of bins and find out how many number of data points falls in the bins for each class. The quantity is measured as given in \eqref{iv_comp}.

\begin{equation} \label{iv_comp}
	\text{IV} = \sum_i \left( \frac{n_p^i}{n_p} - \frac{n_n^i}{n_n} \right) \times \frac{n_p^i/n_p}{n_n^i/n_n}
\end{equation}

The summation in \eqref{iv_comp} is over the bins, $n_p^i$ and $n_n^i$ is the number of data points in positive and negative class respectively in the bin:$i$, and $n_p$ and $n_n$ are the total number of data points in positive and negative class respectively. An \textit{IV} value $>$ 0.5 indicates that the corresponding feature is an extremely strong predictor and values in the interval [0.3, 0.5] indicates strong predictor \cite{shi2020safe}.  In the experiments, the number of bins was chosen as 10 after  removal of data points that are outliers.

\bibliography{example_paper}
\bibliographystyle{IEEEtran}

\end{document}